\documentclass{article}

\usepackage[normalem]{ulem}

\usepackage{templateArxiv}

\usepackage[utf8]{inputenc} 
\usepackage[T1]{fontenc}    
\usepackage{hyperref}       
\usepackage{url}            
\usepackage{booktabs}       
\usepackage{amsfonts}       
\usepackage{nicefrac}       
\usepackage{microtype}      
\usepackage{lipsum}
\usepackage{fancyhdr}       
\usepackage{graphicx}       
\graphicspath{{media/}}     
\usepackage{amsmath}

\usepackage{natbib}
\usepackage{float}

\title{K-Origins: Better Colour Quantification for Neural Networks
}

\author{
  Lewis Mason \\
  MASc Mechanical Engineering\\
  University of British Columbia\\
  Vancouver\\
  \texttt{lewismm@student.ubc.ca} \\
   \And
  Mark Martinez \\
  Department of Chemical and Biological Engineering \\
  University of British Columbia \\
  Vancouver\\
  \texttt{mark.martinez@ubc.ca} \\
}

\begin{document}

\maketitle

\begin{abstract}

K-Origins is a neural network layer designed to improve image-based network performances when learning colour, or intensities, is beneficial. Over 250 encoder-decoder convolutional networks are trained and tested on 16-bit synthetic data, demonstrating that K-Origins improves semantic segmentation accuracy in two scenarios: object detection with low signal-to-noise ratios, and segmenting multiple objects that are identical in shape but vary in colour. K-Origins generates output features from the input features, $\textbf{X}$, by the equation $\textbf{Y}_k = \textbf{X}-\textbf{J}\cdot w_k$ for each trainable parameter $w_k$, where $\textbf{J}$ is a matrix of ones. Additionally, networks with varying receptive fields were trained to determine optimal network depths based on the dimensions of target classes, suggesting that receptive field lengths should exceed object sizes. By ensuring a sufficient receptive field length and incorporating K-Origins, we can achieve better semantic network performance. Examples of these improvements are illustrated in Figure \ref{fig:exampledata}.

\end{abstract}

\section{Introduction}

Semantic segmentation classifies 2D or 3D images on a pixel-by-pixel basis. It is especially valuable for processing large datasets that are impractical to classify manually. In biomedical and materials science, semantic segmentation is particularly useful for two tasks: distinguishing objects from the background and differentiating tracer particles from non-tracer particles.

\begin{figure}
    \centering
    \includegraphics[width=\textwidth]{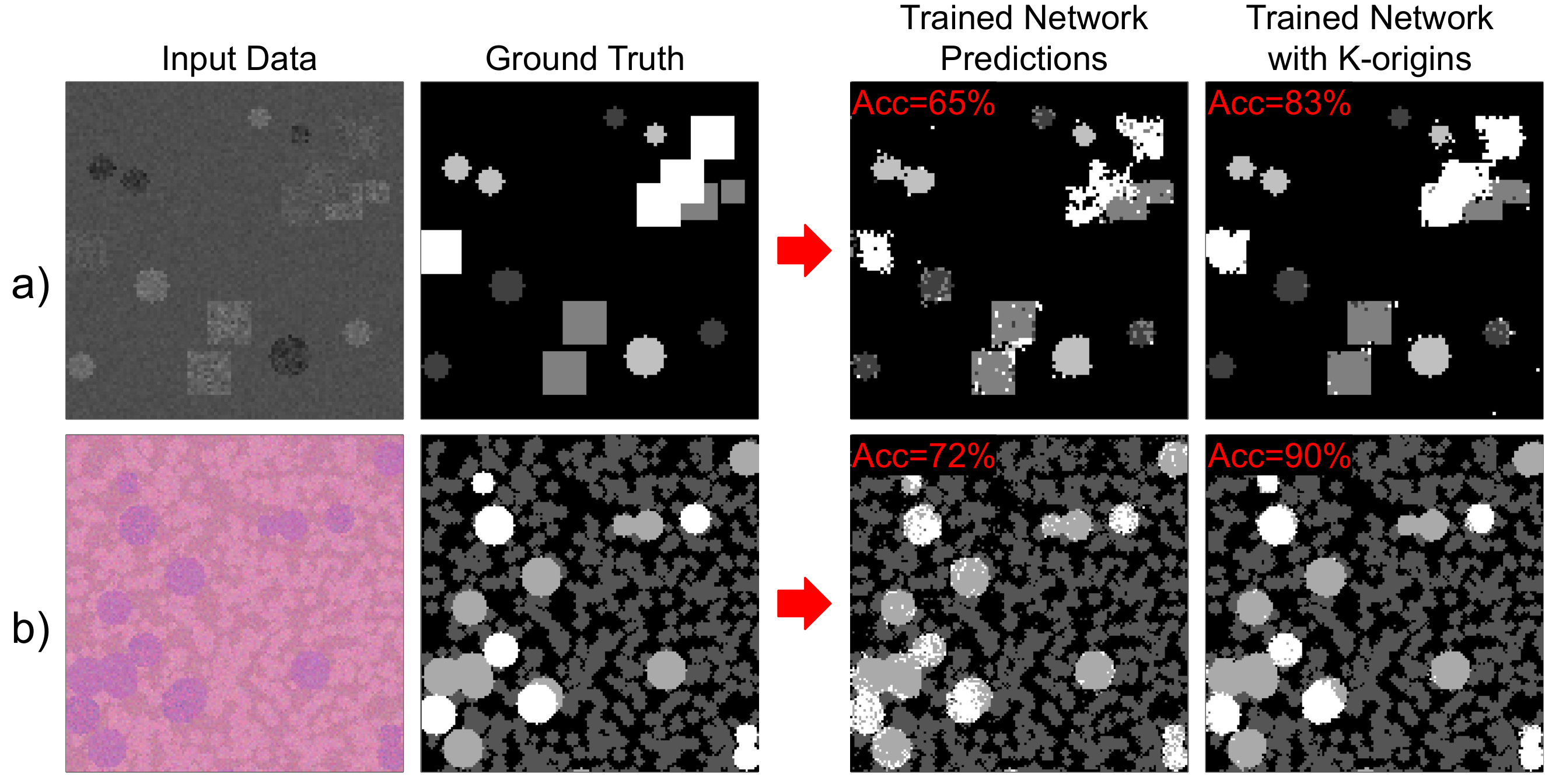}
    \caption[Synthetic data segmentation problem examples]{Two examples of synthetic data segmentation tasks, demonstrating performance improvements from K-Origins. Each colour in the ground truth represents a different class. The neural networks used are discussed later, and only differ by the inclusion of K-Origins. (a) Segmenting multiple noisy grayscale classes from a noisy background after 20 epochs. (b) The "tracer" problem in colour, segmenting nearly identical classes (the largest circles with different greyscale values in the ground truth) with slight variations in colour distributions after 30 epochs.}
    \label{fig:exampledata}
\end{figure}

The first class of problems, object segmentation, involves distinguishing one, or more, target classes from the background. Examples include \cite{BANIK2020113211} where white blood cell nuclei are segmented, \cite{LARSSON2018465} where abdominal organs and regions of interest are segmented, and \cite{furat2019machine} where X-ray tomography images of materials such as liquid-solid composites and ore-particles are segmented. This segmentation problem is prevalent in engineering, biomedical research, and materials sciences.

On the other hand, tracer segmentation involves distinguishing objects that are nearly identical in shape, but vary by colour or intensity. An example of this problem is segmenting X-ray images where contrast enhancing agents have been used (\cite{li2014contrast} and \cite{de2015utilization}). Another example is cancer cell segmentation in pathology slides, like in \cite{WANG20191686}, where cancerous cells can vary from normal cells by colour. The tracer segmentation problem is also relevant for datasets that produces a large number of false positives during segmentation.

Convolutional neural networks (CNNs) have shown to be very good at semantic segmentation and one of the most popular architecture styles for this task is the encoder-decoder network. This architecture makes predictions by combining low-level and high-level image features, effectively integrating information from various fields of view to achieve optimal results. The encoder-decoder network was mainly popularized by U-Net (\cite{ronneberger2015unet}) which has been cited over 89,000 times. Because of its wide spread use, U-Net serves as a basic blueprint for the architectures used in this paper.

The receptive field (RF) is a key characteristic of CNNs. It represents the network's 2D or 3D field of view, indicating how much of the input image is used at each feature layer. To differentiate the RF, an area or volume, from its side length, we refer to the side length as the receptive field length (RFL). The RFL can be calculated for one side using the recursive equation from \cite{araujo2019computing}:

\begin{equation}
    r_{l-1} = s_l \cdot r_l + (k_l - s_l)
    \label{eq:rfl}
\end{equation}

To determine the RFL before a layer in the network ($r_{l-1}$) given the RFL after that layer ($r_l$), use the layer's stride ($s_l$) and kernel size ($k_l$) in Equation \ref{eq:rfl}. For semantic segmentation, start at the deepest set of features with an RFL of one ($r_{l=end} = 1\ \mathrm{pixel}$) and work backwards to determine the RFL at each feature layer. 

The RFL at the beginning of semantic networks can then be thought of as the side length of the area or volume used for a single pixel's prediction. For example, a 2D semantic network with an RFL of 11 uses an 11x11 pixel area to generate the features used for classifying the central pixel. If the RFL is symmetrical in all directions, it only needs to be calculated for one dimension.

RFs have been extensively studied in various articles, with some focusing on determining optimal sizes and the effective RFLs in networks: \cite{luo2016understanding, liu2018understanding, gabbasov2020influence}. However, many studies use complex datasets, making it difficult to generalize the findings.

Increasing the complexity of neural networks often improves training accuracy but results in longer training times and higher hardware costs. With millions of trainable parameters, it also becomes difficult to understand what the network is learning. It would be beneficial if networks could be made smaller and more efficient without hurting their performance.

This paper aims to reduce neural network complexity by building architectures from the ground up with synthetic data, ensuring that the correct properties, such as colour and shape, are learned effectively. It deviates from the standard research structure, as it addresses no obvious deficiencies in CNN research. Instead, the work is motivated by testing neural networks on simple datasets to identify shortcomings which can then be resolved. We also look at how the RFL affects results using this simple dataset. Overall, our goal is to decrease network complexity without negatively affecting the results.

In Section \ref{sec:syntheticdata}, we discuss the data generation process for all trials. The motivating case for this study is presented in Section \ref{sec:motivation}, demonstrating that a CNN can struggle with simple object detection. In Section \ref{sec:metrics} we introduce some additional background material that is relevant for quantifying results. In Section \ref{sec:clayer}, we introduce K-Origins, a layer designed to help neural networks quantify colours and intensity magnitudes. Section \ref{sec:rfl} demonstrates that the motivating case can either be solved by using K-Origins or by increasing the depth and complexity of the network. Finally, in Section \ref{sec:mixed}, we test the limits of segmentation across a range of colour distributions for two types of problems: object detection and tracer segmentation.

\section{Methods}\label{sec:methods}

\subsection{Synthetic data}\label{sec:syntheticdata}

Greyscale synthetic data is generated with a 16-bit colour channel for various test cases. This data contains a background with randomly placed squares, and the number of squares varies to ensure that the background remains visible. For each trial, 400 synthetic images with the dimensions 200x200 are created for training, and an additional set is used for testing. Square side lengths and class intensity distributions vary between trials. By using squares, which are simpler shapes and are easier to interpret, we can better assess the impact of K-Origins.

\begin{figure}
    \centering
    \includegraphics[width=\textwidth]{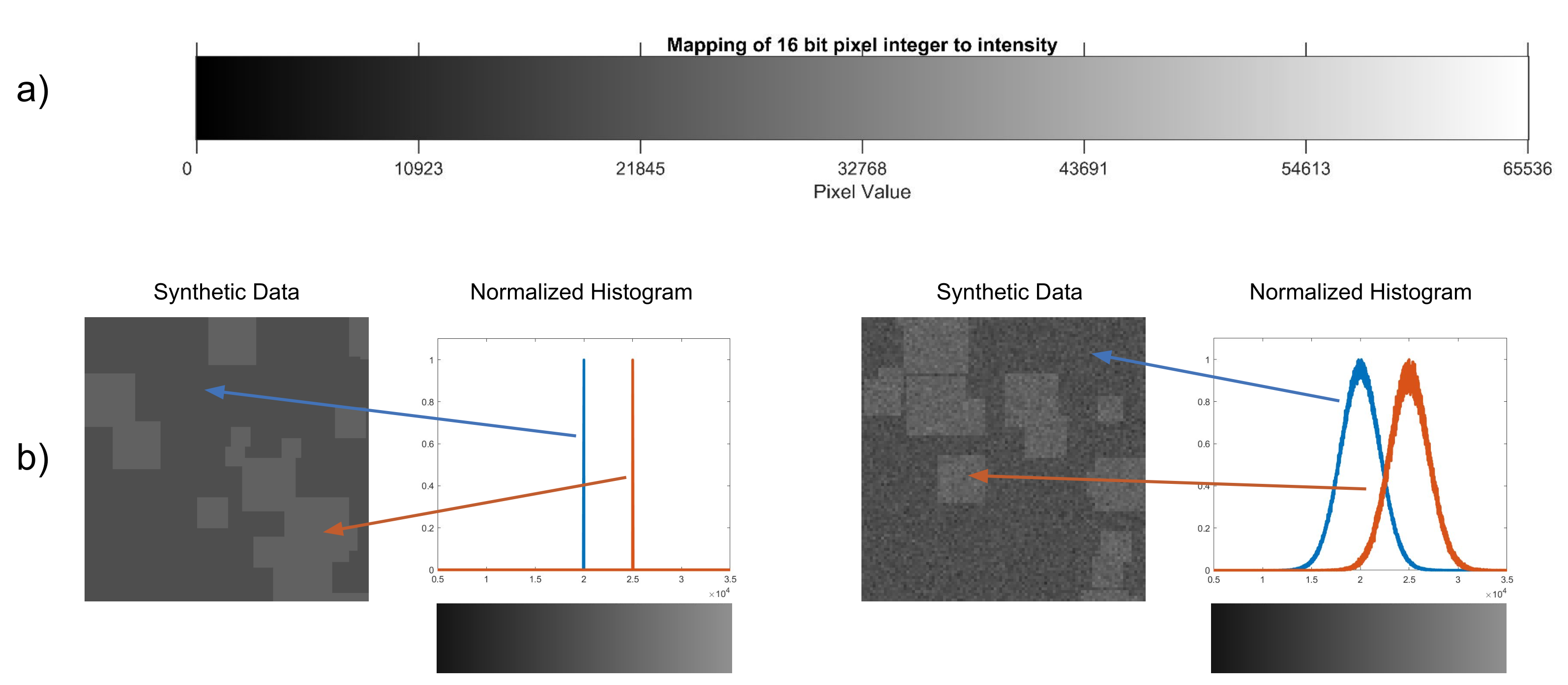}
    \caption[Integer-greyscale mapping]{(a) Mapping from 16-bit pixel integer values to grayscale intensity. (b) Example of synthetic data used in this paper with corresponding histograms for both noise-free and Gaussian noise cases.}
    \label{fig:colourexplanation}
\end{figure}

For greyscale data, a pixel's colour is represented by a single integer value. For 16-bit data, as used in this paper, the values range from 0 (pure black) to 65535 (pure white), with various shades of grey in between. In this work, a class's colour is represented by its intensity mean ($\mu_{i}$) and the standard deviation of added Gaussian noise ($\sigma_{i}$). Figure \ref{fig:colourexplanation} shows the integer-intensity mapping and provides examples of the synthetic data used in this paper. Data intensity distributions are illustrated using normalized histograms ($data = data/max(data)$).

\subsection{Motivating Case: Network Failure}\label{sec:motivation}

\begin{figure}
    \centering
    \includegraphics[width=\textwidth]{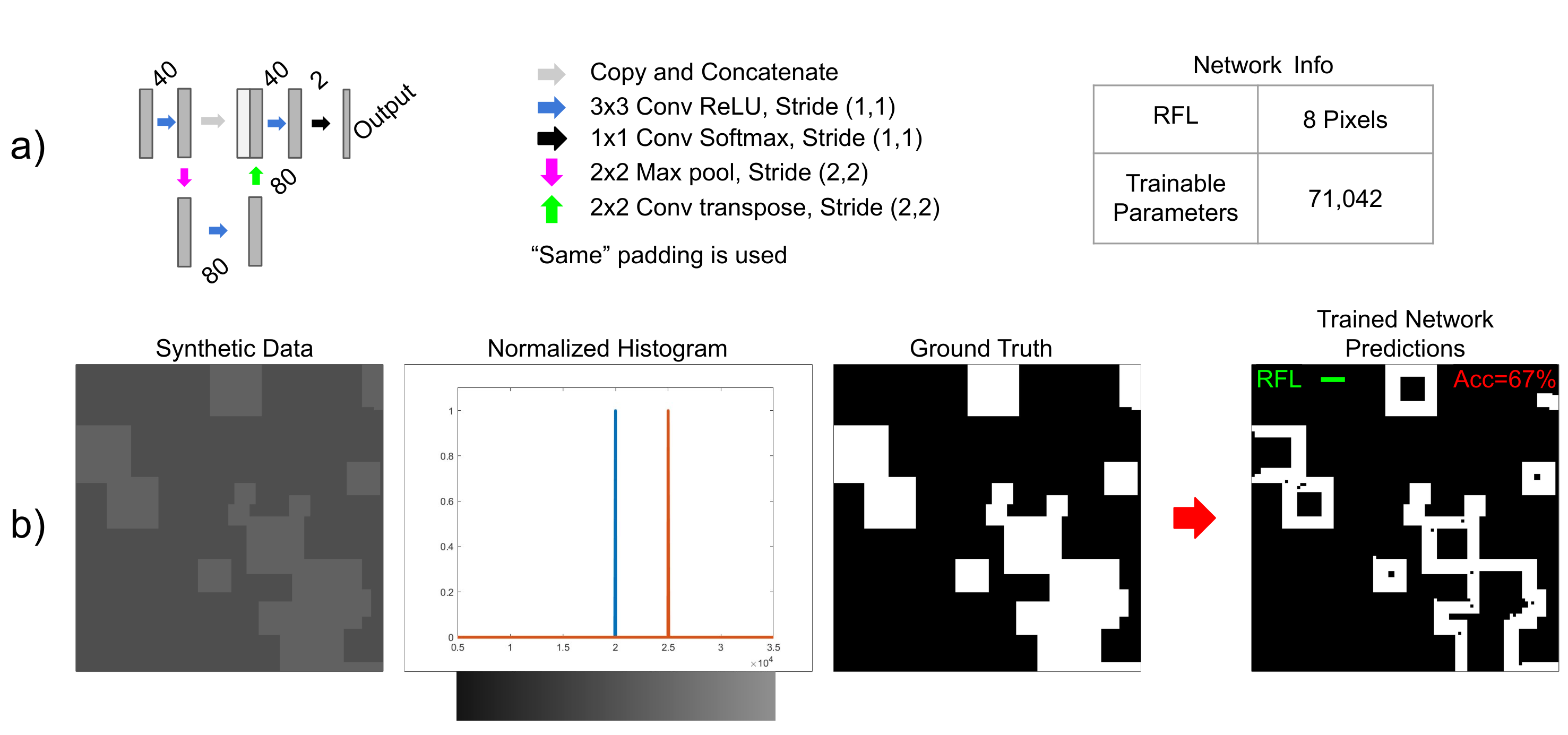}
    \caption[Small CNN failure on trivial data]{Small encoder-decoder network predicting synthetic data with square side lengths ranging from 5 to 20 pixels. (a) Network architecture. (b) A 67\% validation accuracy at steady state after 77 epochs with a learning rate of 1E-3. This shows input data, the histogram, the ground truth, and the network prediction. The network struggles with colour magnitudes, correctly classifying only up to 4-5 pixels from the object borders, indicating reliance on colour gradients rather than colour magnitudes for classification.}
    \label{fig:motivation}
\end{figure}

The motivation for K-Origins and this work is shown in Figure \ref{fig:motivation}, where a small encoder-decoder network fails to classify noiseless squares from the background. The network lacks an understanding of colour magnitude; if it could recognize the lighter gray squares against the darker gray background, the task would be simple. However, the network does not directly leverage the 16-bit values—the greyness—of the squares in its predictions. For example, a straightforward solution to this problem is to compare a pixel's integer value to 25000 (the squares' colour) and classify it as a square if it matches, or as background if it does not. Despite having over 70,000 trainable parameters, the network fails to learn this behavior. 

Moreover, the network can only correctly classify squares within 4 to 5 pixels from the object border. This suggests the network detects gradients rather than colour magnitudes and does so over a specific length. Convolutions are known for detecting gradient-related behavior so this is almost expected, but it would be highly beneficial if the non-linearity of neural networks could be used to leverage colour magnitudes more directly.

In Figure \ref{fig:motivation}, the network's RFL is calculated by setting the bottom-right feature (the deepest point) to $r_{l=end} = 1$ pixel and recursively determining the RFL at previous layers. Using Equation \ref{eq:rfl}, we calculate that the motivating network has an RFL of 8 pixels, which is twice the distance that gets correctly classified from the object border, plus or minus one pixel. Being twice the correct prediction distance should be expected because the deepest features in that network can "see" about 4-5 pixels on either side of the pixel it wishes to classify. We hypothesize that this network classifies pixels by detecting a square edge in any direction; if no edge is detected within the RF, the pixel is classified as background.

Figure \ref{fig:motivation} shows that the network struggles to classify pixels far from the object border and that it also fails to understand intensity magnitudes. We will address both of these issues separately and will use greyscale data because the single channel results extend to additional colour channels (RGB).

\subsection{Metrics}\label{sec:metrics}

In this section we introduce important equations and data properties that will be used throughout the rest of the paper.

To quantify the distance between intensity distributions, we use the Hellinger distance for two classes represented by Gaussian probability density functions (PDFs). The HD for two Gaussian distributions is given by:

\begin{equation}\label{eqn:hellinger}
    \mathrm{HD}(\mathcal{N}(\mu_1,\sigma_1),\mathcal{N}(\mu_2,\sigma_2)) = \sqrt{1 - \sqrt{\frac{2\sigma_1 \sigma_2}{\sigma_1^2 + \sigma_2^2}}exp\left( -\frac{1}{4}\frac{(\mu_1 - \mu_2)^2}{\sigma_1^2 + \sigma_2^2}\right)}
\end{equation}

where $\mathcal{N}$ is a normal distribution with means $\mu_1$ or $\mu_2$ and standard deviations $\sigma_1$ or $\sigma_2$ respectively (\cite{ding2023empirical}). This equation produces a value between 0 and 1, where an HD of 0 indicates identical distributions, and an HD of 1 indicates completely distinguishable distributions.

Next, we introduce a modified accuracy metric to address class imbalance. In this paper, the background (class zero) is so large that it inflates the accuracy score. To counter this, we exclude the background class from all accuracy calculations. The resulting modified accuracy is given by:

\begin{equation}\label{eqn:customacc}
    \mathrm{MAcc} =  \frac{1}{C-1}\sum_{i\neq \mathrm{background}}^{C}\frac{TP_{i}}{TP_{i} + FP_{i} + FN_{i}}
\end{equation}

where $MAcc$ is the custom accuracy with background bias removed, $C$ is the total number of classes including the background, $TP_i$ represents true positives, $FP_i$ false positives, and $FN_i$ false negatives for class $i$. A target class is any class that is not background ($i \neq background$). This turns out to be the Jaccard index (\cite{taha2015metrics}) for multiple classes and throughout this paper all mentions of accuracy are referring to MAcc.

\subsection{K-Origins Layer: The Colour Solution}\label{sec:clayer}

We first develop a layer to help networks quantify colour magnitudes. To the best of the authors' knowledge at the time of writing, this approach is unique. Given features $\textbf{X}$, a K-Origins layer with $K$ trainable weights produces output features for each trainable weight $w_k \in [w_1,w_2,...,w_K]$, as follows:

\begin{equation}
    \textbf{Y}_k=\textbf{X}-\textbf{J} \cdot w_k
\end{equation} 

where $\textbf{Y}_k$ is the output given from a single weight $w_k$, and $\textbf{J}$ is a matrix of ones matching the dimensions of $\textbf{X}$. This layer produces $K$ copies of the input image, each with a different scalar subtracted from it, resulting in $K$ images with different origins. All values less than the weight $w_k$, or origin, become negative in $\textbf{Y}_k$ and all data greater than $w_k$ stays positive. 

For 2D and 3D image data there is normally one origin (zero), making all data positive relative to it. If we immediately use K-Origins on this input data, then future layers such as convolutions can use the sign changes to determine the relative intensity locations for each pixel. Similar behaviour can be done for deeper features in a network. For the first K-Origins layer the weights $w_k$ must match the data's (or features) order of magnitude. For un-signed 16-bit data we see $w_k \in [0, 65535]$ for the first layer, requiring learning rates of 1-100 for significant parameter changes during training.

\begin{figure}
    \centering
       \includegraphics[width=\textwidth]{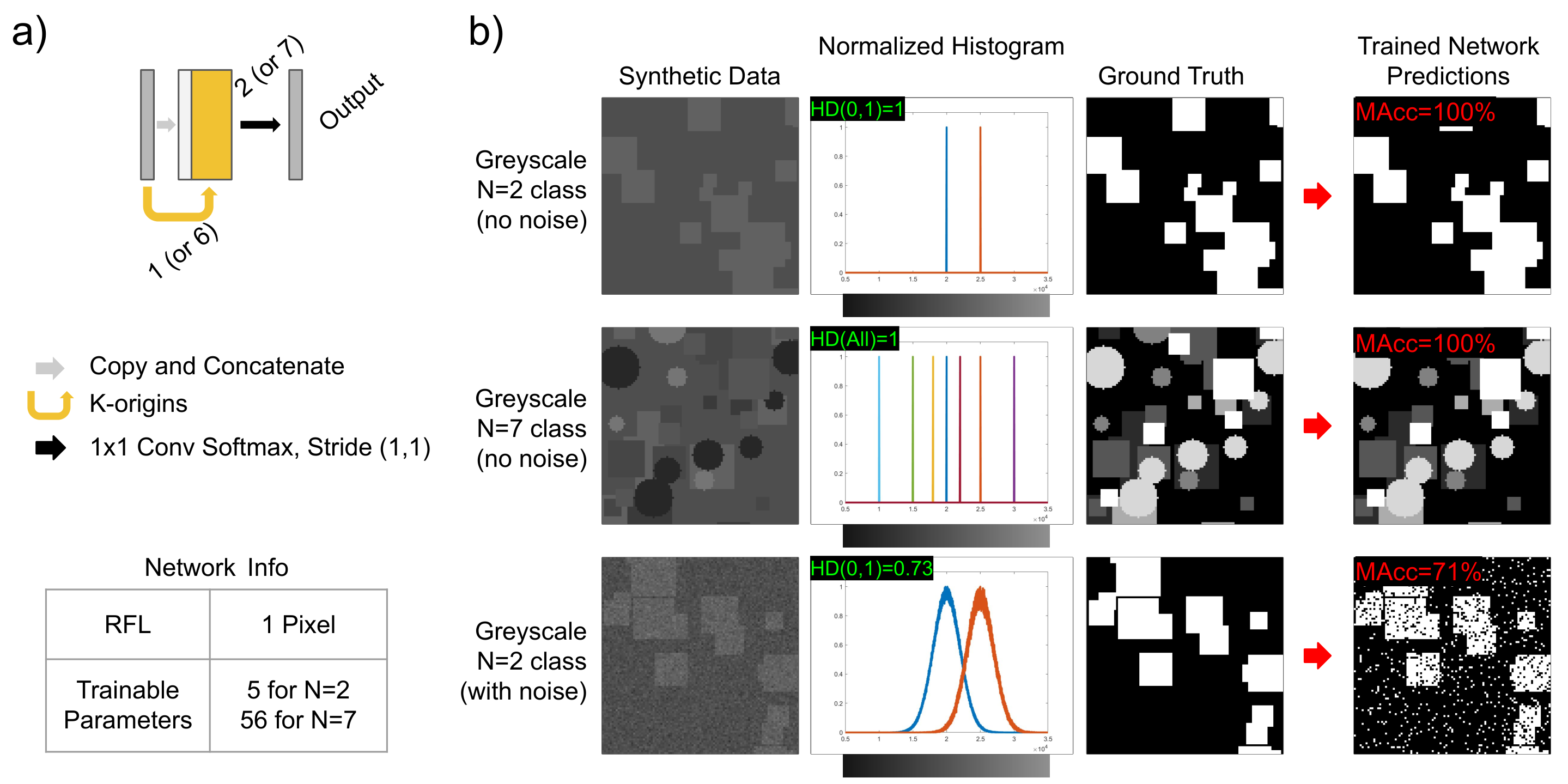}
    \caption[Basic K-Origins Network Example]{(a) Very small colour network architecture and additional parameters. (b) Colour network results on the motivating case that the smaller encoder-decoder network failed to solve, the motivating case with additional classes, and a failure case with introduced noise, where the Hellinger distance is no longer unity between classes.}
    \label{fig:clayer}
\end{figure}

Figure \ref{fig:clayer} shows a small network that takes an input image and concatenates it with the output of a K-Origins layer with one weight, $w_1$. Concatenating the output of K-Origins with the input provides stable reference features for the rest of the network, which is essential for convergence as $\textbf{Y}_k$ constantly changes. The network then applies a softmax-activated 1x1 convolution with a learning rate of 1E-3 for pixel-wise predictions. Because this network has an RFL of one pixel, it can only use information from a single pixel for its predictions, extracting no spatial information.

The weight $w_1$ was initialized at 50000 with a learning rate of 100 and ended at $w_{1,final} = 20200$ after 33 epochs. This final value lies between the intensity values of the two classes, $\mu_0 =20000$ and $\mu_1 = 25000$. This small network with only 5 trainable parameters achieved 100\% accuracy segmenting the case from Figure \ref{fig:motivation}, whereas the encoder-decoder network with 71,042 trainable parameters achieved only a 67\% accuracy. This small network was also tested with more weights on a 7-class case and achieved 100\% accuracy. However, accuracy decreased when the class intensity distributions had an HD less than unity, suggesting that a combination of K-Origins and shape recognition would perform better.

Because supervised learning problems have ground truths, K-Origins weights can be initialized based on known class distributions with learning rates of zero, or near zero. For example, in the first problem of Figure \ref{fig:clayer}b, initializing the K-Origins weight as $w_1=22500$, right between both classes, achieves 100\% accuracy in just one epoch. This technique is used later in this article by "clamping" distributions, where a weight is placed above and below the known distribution of a target class to clamp those intensities.
 
Convolutional neural network layers are generally defined as:

\begin{equation}
    \textbf{Y} = f(\textbf{X}\ast \textbf{c} + b)
\end{equation}

where $\textbf{X}\ast \textbf{c}$ is the convolution of features $\textbf{X}$ with kernel $\textbf{c}$, $f(z)$ is the activation function, and $b$ is the bias. Often for semantic segmentation networks the reluctance (ReLU) activation function is used ($f(z)=ReLU(z)$), such as in \cite{ronneberger2015unet}. ReLU forces negative numbers to be near zero and positive, making it hard for a network to learn the behavior of K-Origins without directly implementing it. While a convolutional network could theoretically learn similar behavior, it would be challenging. 

Next we look at setting various neural network depths and compare accuracies with and without K-Origins for a range of RFL's.

\subsection{RFL's: Length Scale Solution}\label{sec:rfl}

\begin{figure}
    \centering
    \includegraphics[width=\textwidth]{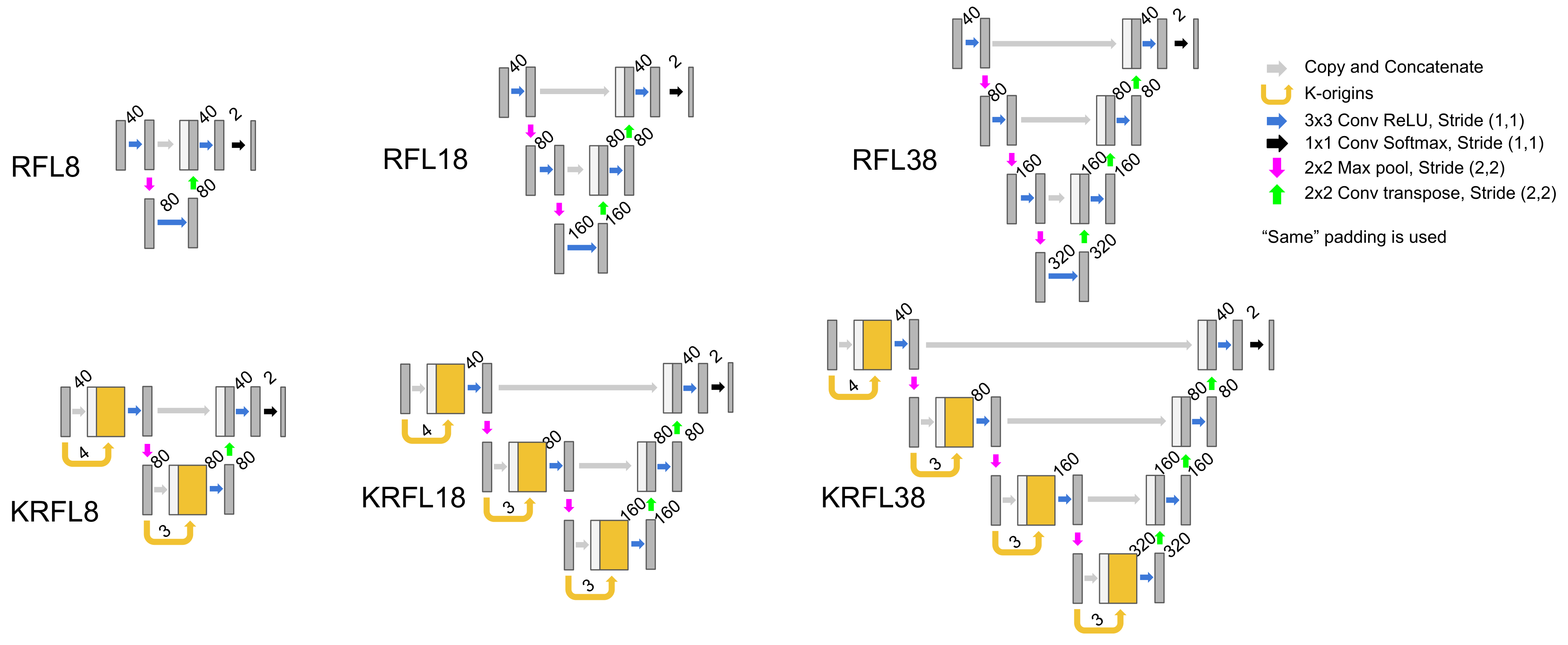}
    \caption[Set of architectures used for RFL tests]{Set of neural networks used for receptive field length tests with and without K-Origins. Networks RFL8, RFL18, and RFL38 are miniature U-Net based architectures, differing primarily in their receptive field length. Networks KRFL8, KRFL18, and KRFL38 are identical but include K-Origins at every depth.}
    \label{fig:rflarchitectures}
\end{figure}

In Figure \ref{fig:motivation}, the network fails for larger objects. In this section, we investigate the required RFL for various object sizes. We use a set of small encoder-decoder networks, shown in Figure \ref{fig:rflarchitectures}, with additional details in Table \ref{tab:architectures}. The six architectures used are RFL8, RFL18, RFL38, KRFL8, KRFL18, and KRFL38, where "KRFLX" refers to an identical architecture to "RFLX" with K-Origins. All networks in Table \ref{tab:architectures} use "same" padding, where applicable, to prevent cropping. We hypothesize that the RFL should be larger than the dominant length scale, or the minimum length required to differentiate two objects.

\begin{table}
    \centering
    \begin{tabular}{ccccc}
    \toprule
         Network Name &  RFL & Depth & \# Parameters & Section Used\\
         \midrule
         \midrule
         RFL8 &  8 & II & 71,042  & \ref{sec:rfl}\\
         RFL18 &  18 & III &  352,962 & \ref{sec:rfl}\\
         RFL38 &  38 & IV &  1,480,002 & \ref{sec:rfl}\\
         \midrule
         KRFL8 &  8 & II &  187,846 & \ref{sec:rfl}\\
         KRFL18 &  18 & III & 930,886 & \ref{sec:rfl}\\
         KRFL38 &  38 & IV &  3,901,766 & \ref{sec:rfl}\\
         \midrule
         2 Class RFL14 &  14 & II &  330,522 & \ref{sec:mixedonetarget}\\
         2 Class KRFL14 &  14 & II & 274,406 &  \ref{sec:mixedonetarget} \\
         \midrule
         3 Class RFL14 &  14 & II &  445,843  & \ref{sec:mixedtwotarget} \\
         3 Class KRFL14 &  14 & II &  275,169 & \ref{sec:mixedtwotarget}\\
         \bottomrule

    \end{tabular}
    \caption[Table of synthetic network architectures and additional information]{Table of all network architectures based on the U-Net architecture explored in this paper.}
    \label{tab:architectures}
\end{table}

We first train the six networks on noiseless data ($\mu_0 = 20000,\ \mu_1 = 25000,\ \sigma_0=\sigma_1 = 0$) containing squares with a side length of 25 pixels, similar to the scenarios in Figures \ref{fig:motivation} and \ref{fig:clayer}. We also train on noisy data ($\mu_0 = 20000,\ \mu_1 = 25000,\ \sigma_0=\sigma_1 = 2000$) to simulate the failure case in Figure \ref{fig:clayer}. This shows us the effect of increasing the RFL for a fixed object size. 

Training runs for 10 epochs with a batch size of 3. Learning rates are set to 1E-3 for convolution layers and 100 for K-Origins layers. The highest-level K-Origins weights are initialized by placing a weight two standard deviations above and below the intensity mean for each class ($w_{i1,i2} = \mu_i \pm 2\sigma_i$). This effectively clamps each class's intensity distribution with two K-Origins parameters. For the noiseless case this corresponds to ${w_i}=\{20000,20000,25000,25000\}$, and for the noisy case, ${w_i}=\{16000,24000,21000,29000\}$. All other K-Origins layers have three weights initialized from Gaussian random variables with $\mu = 20000$ and $\sigma=5000$. 

\begin{figure}
    \centering
    \includegraphics[width=\textwidth]{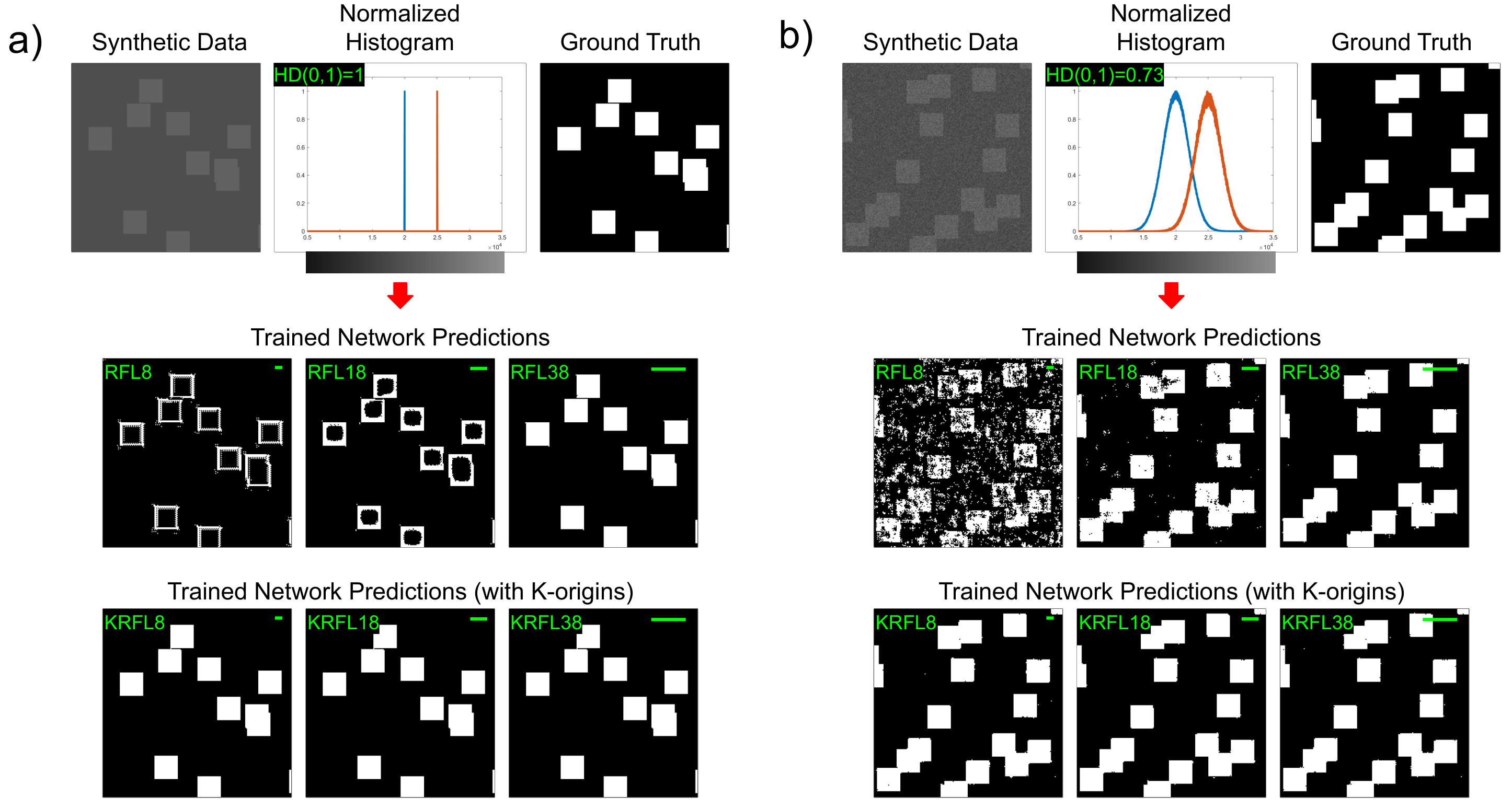}
    \caption[Varying network RFLs for fixed object sizes results]{Example training results for varying RFLs on squares with a side length of 25 pixels, shown for networks with and without the K-Origins layer. Actual RFL lengths are indicated by a green line in the top right of each prediction image. (a) No noise case where $\mu_0 = 20000$ and $\mu_1 = 25000$. (b) Case with Gaussian noise added, where $\sigma_0 = \sigma_1 = 2000$. For networks without K-Origins, accuracy increases as RFL approaches and exceeds the dominant object length scale. All networks with K-Origins achieve a high accuracy regardless of RFL, as they can learn and use the colours of the target objects for predictions.}
    \label{fig:rflresultsexample}
\end{figure}

The results from these trials are shown in Figure \ref{fig:rflresultsexample}. Networks without K-Origins increase in accuracy as the RFL approaches the object length, achieving high accuracies when the RFL exceeds the object length.  In contrast, networks with K-Origins achieve near-perfect validation accuracy regardless of their RFL, demonstrating a more efficient solution. Achieving a near-perfect accuracy without K-Origins requires about 1.4 million trainable parameters, while using K-Origins achieves the same accuracy with only 187,000 trainable parameters. Additionally, an even smaller network with K-Origins could be possible, as this test did not determine a network size lower bound.

\begin{figure}
    \centering
    \includegraphics[width=\textwidth]{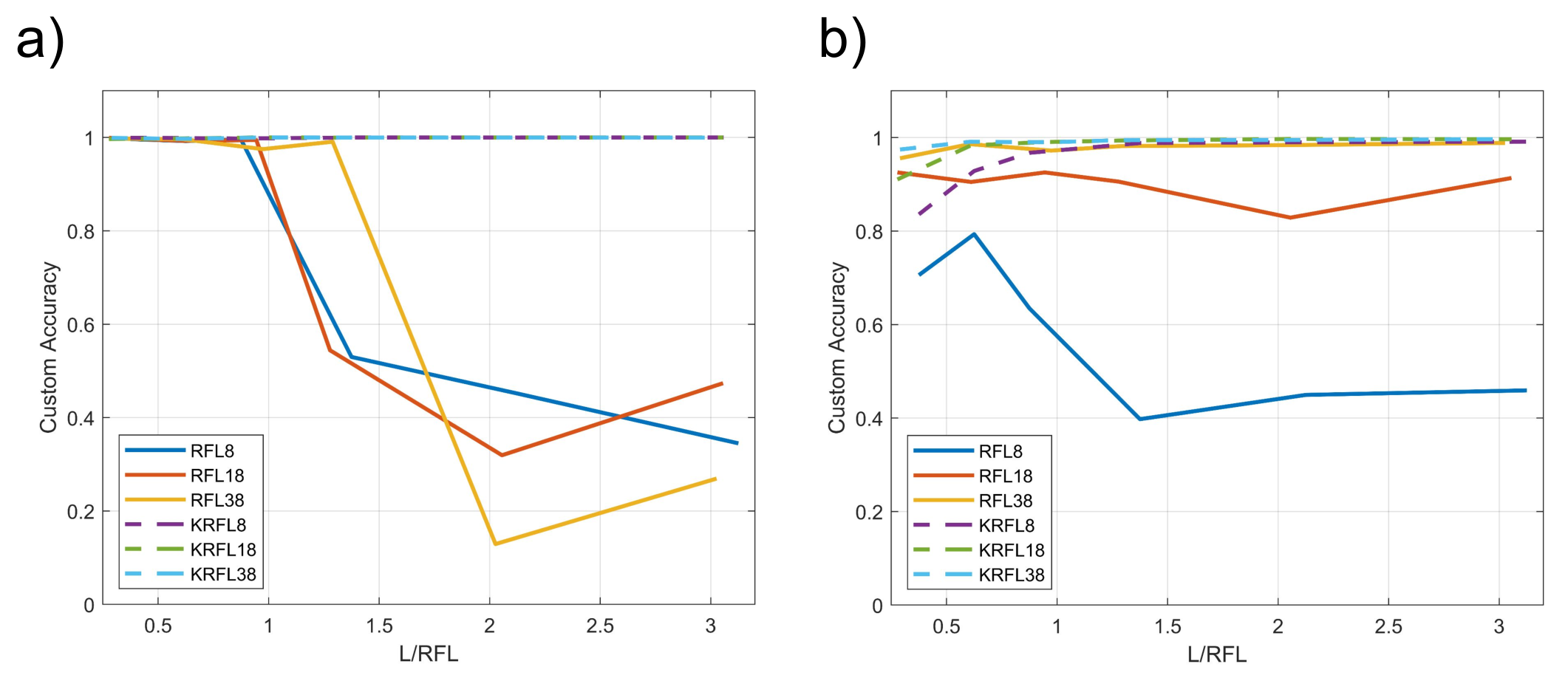}
    \caption[Validation results for a sweep of RFLs and object sizes]{Validation results for varying-sized squares with RFLs of 8, 18, and 38 for networks with and without K-Origins. The object size to RFL ratios investigated are $L/RFL\approx \{0.3,0.6,0.95,1.3,2,3\}$, with side lengths rounded to whole numbers. (a) No-noise case where $\mu_0 = 20000$ and $\mu_1 = 25000$. (b) Case with Gaussian noise added, where $\sigma_0 = \sigma_1 = 2000$. Networks with K-Origins outperform those without it due to a better usage of object colours and intensities. Networks without K-Origins perform best when the target object size is less than the RFL.}
    \label{fig:rflresults}
\end{figure}

Next, we perform a sweep of square side lengths to RFL ratios, $L/RFL$, for the six networks using the same training parameters as before. This is done with and without noise. For each RFL, we examine $L/RFL\approx \{0.3,0.6,0.95,1.3,2,3\}$. These fractions are approximated since side lengths may be rounded. The summary of these tests is shown in Figure \ref{fig:rflresults} and in almost every case, using K-Origins increases accuracy. We also observe that accuracy decreases when $L/RFL$ is small. This is because a small $L/RFL$ results in very small squares, making segmentation difficult in noisy conditions regardless of the architecture used. All numerical results are found in Appendix \ref{app:data}.

In almost every case, networks with K-Origins outperform those without it. For this problem, KRFL8, KRFL18, and KRFL38 achieved nearly 100\% accuracy in about 3 epochs, compared to the 10 epochs for their RFLX counterparts. While it might be argued that this is due to the high learning rate of K-Origin layers, training with a learning rate of zero for K-Origins produces similar results with the same initialization. The use of K-Origins may enable smaller and more efficient networks without sacrificing performance.

Networks without K-Origins (RFLX) succeed when the RFL is larger than the object size, which aligns with the preference for very deep networks in most research. These networks also seem to perform better on noisy data than on noiseless data. 

So far we have demonstrated a solution to the motivational problem (Figure \ref{fig:motivation}) using both network length scales (ensuring sufficient RFL) and intensity quantification (K-Origins) for a noisy and noiseless case. In the noiseless case we set $\Delta \mu = 5000$ and $\sigma = 0$, giving a unity HD. In the noisy case we set $\Delta \mu$ is the same, but $\sigma = 2000$ resulting in an HD of 0.73. The next logical step is to sweep across various HDs by adjusting $\Delta \mu$ and $\Delta \sigma$ to determine the effectiveness of K-Origins for different intensity distributions.

\subsection{Mixed Solution for a Range of Intensity Distributions}\label{sec:mixed}

\begin{figure}
    \centering
    \includegraphics[width=0.7\textwidth]{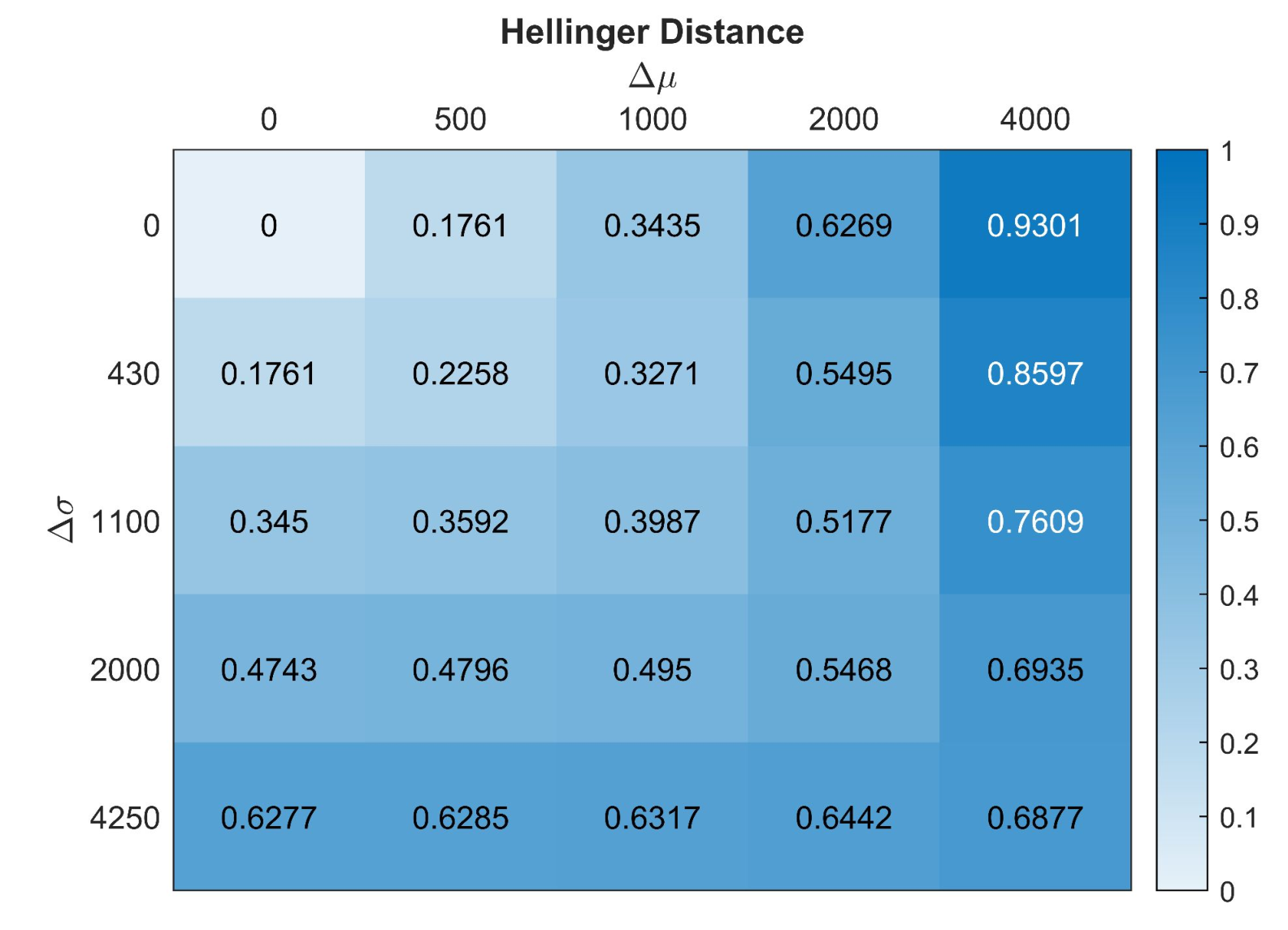}
    \caption[Heatmap of Hellinger distances between two Gaussian distributions]{Heatmap of Hellinger distances for various $\Delta \mu$ and $\Delta \sigma$ between two Gaussian distributions. The reference class distribution has $\mu = 20000$ and $\sigma = 1000$. The X-axis represents the change in the mean of the second distribution relative to the first. The Y-axis represents the change in the noise standard deviation of the second class relative to the first. Values indicate the Hellinger distance for the given parameters, with 0 representing identical intensity distributions and 1 representing completely distinct distributions.}
    \label{fig:hellingerdistance}
\end{figure}

In this section we explore how changing the HD affects segmentation by varying $\Delta \mu$ and $\Delta \sigma$. Setting the first class to $\mu=20000$ and $\sigma = 1000$, we produce the HD heatmap shown in Figure \ref{fig:hellingerdistance}. This heatmap will be used to determine HDs for the upcoming trials, allowing us to compare the accuracy of each trial with the HD.

\begin{figure}
    \centering
    \includegraphics[width=\textwidth]{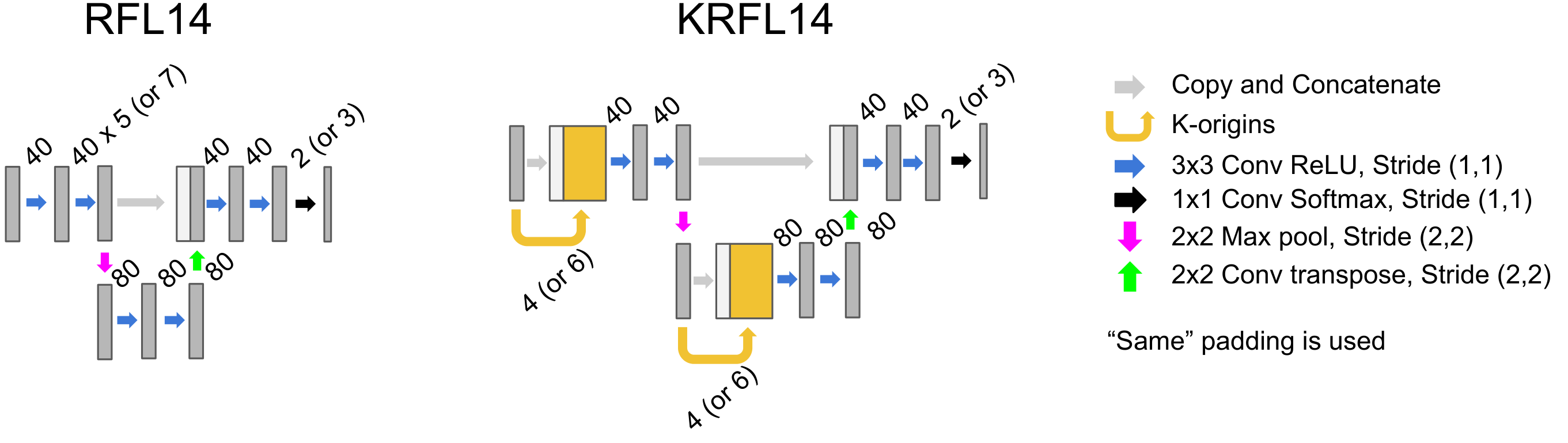}
    \caption[Synthetic networks for a sweep of Hellinger distances]{The two neural networks used for intensity distribution sweeps in the object detection and the tracer segmentation problems. RFL14 is a standard encoder-decoder network, while KRFL14 includes K-Origins. The first convolutional layer in the second level is set to have a similar number of input features for both networks ($40\times 5 = 40\times\left(4+1\right)$).}
    \label{fig:hellingerarchitectures}
\end{figure}

This comparison is done for two network architectures: RFL14 and KRFL14, shown in Figure \ref{fig:hellingerarchitectures} with additional parameters listed in Table \ref{tab:architectures}. The key difference is the inclusion of K-Origins in KRFL14. 

We consider two scenarios: a single target class on a noisy background and two target classes on a noisy background. These problems have two and three output classes, respectively, with the background considered a class. Noise parameter sweeps are performed for two object sizes: squares with side lengths randomized between 6 to 12 pixels ($L<RLF$) and 20 to 30 pixels ($L>RFL$). This results in a total of four cases: object detection with $L<RFL$ and $L>RFL$, and the tracer problem with $L<RLF$ and $L>RFL$.

Networks are trained for 10 epochs with a batch size of 3. Learning rates are set to 1E-4 for convolution layers and 100 for K-Origins layers. K-Origins initialization follows the same method as in Section \ref{sec:rfl} and this time KRFL14 has fewer parameters than RFL14. There are 50 randomly placed squares for $L<RFL$ and 25 for $L>RFL$. All numerical results are found in Appendix \ref{app:data}.

\subsubsection{One Target Class: Object Detection}\label{sec:mixedonetarget}

In this section, we segment a single target class (squares) from a background with an intensity mean of $\mu_0 = 20000$ and noise with a standard deviation of $\sigma_0 = 1000$. We vary the target class mean and standard deviation, where $\Delta \mu = \mu_1-\mu_0$ and $\Delta \sigma = \sigma_1 - \sigma_0$. For each mean and standard deviation, we train both RFL14 and KRFL14 and save the validation accuracies in a heatmap. The x-axis represents the change in mean ($\Delta \mu$), and the y-axis represents the change in standard deviation ($\Delta \sigma$).

\begin{figure}[ht]
    \centering
    \includegraphics[width=\textwidth]{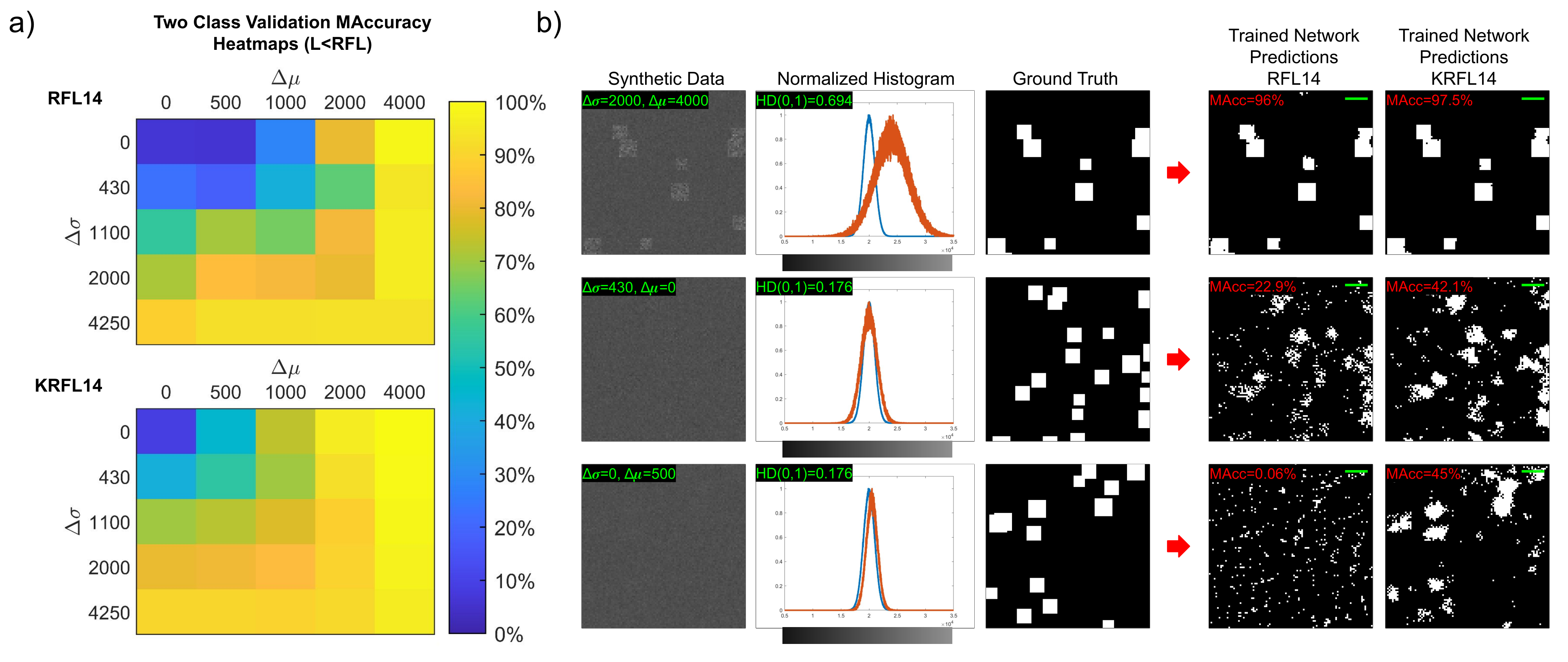}
    \caption[One target class $L<RFL$ results]{RFL14 and KRFL14 validation results for object detection with $L<RFL$ synthetic data. (a) Segmentation validation results visualized in a heatmap for various Hellinger distance variations, represented by shifting means and standard deviations. (b) Three example cases: an easy-to-understand one followed by the two hardest cases. Despite the hardest cases being near indistinguishable, the network still achieves usable segmentation results.}
    \label{fig:twoclassless}
\end{figure}

\begin{figure}[ht]
    \centering
    \includegraphics[width=\textwidth]{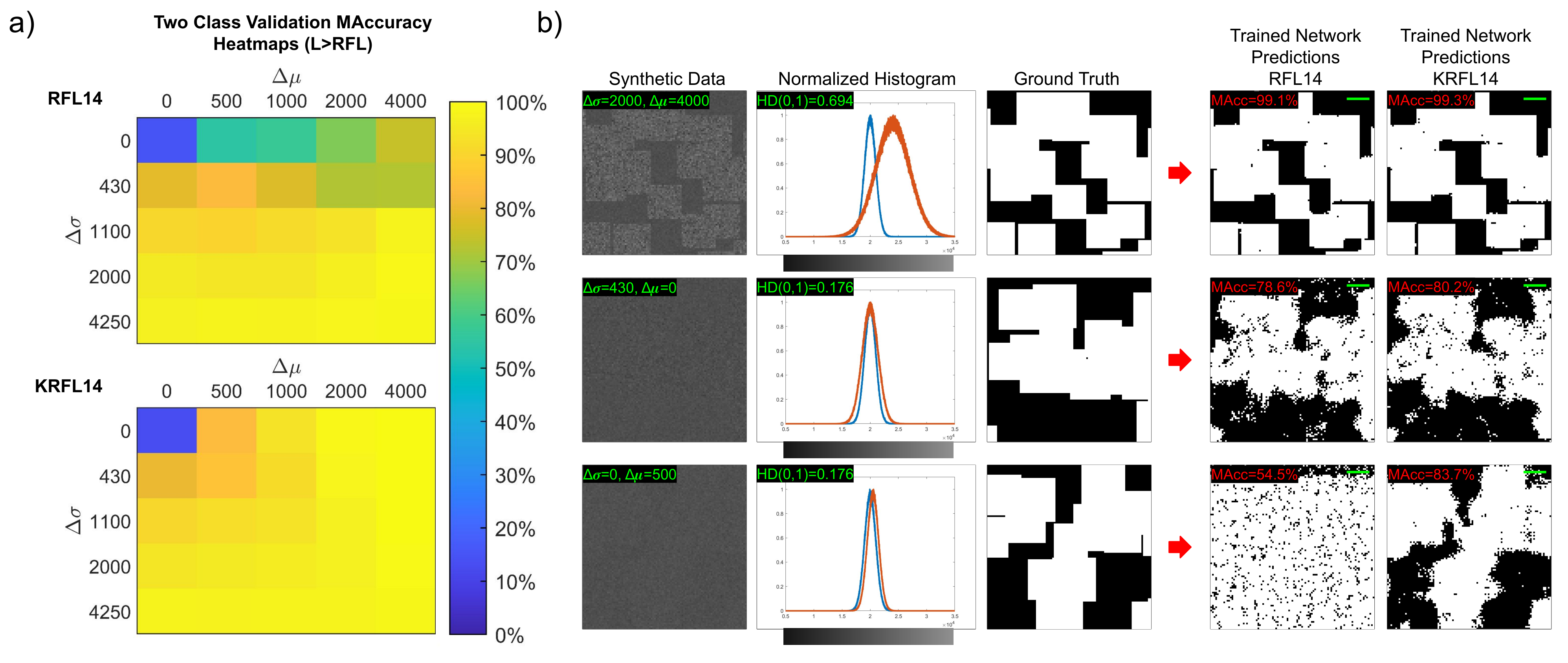}
    \caption[One target class $L>RFL$ results]{RFL14 and KRFL14 validation results for object detection with $L>RFL$ synthetic data. (a) Segmentation validation results visualized in a heatmap for various Hellinger distance variations, represented by shifting means and standard deviations. (b) Three example cases: one easy-to-understand example followed by the two hardest cases.}
    \label{fig:twoclassmore}
\end{figure}

Figure \ref{fig:twoclassless} shows results for $L < RLF $, and Figure \ref{fig:twoclassmore} shows results for $L > RFL$. In both figures, part (a) presents the heatmap with training results for each network. These accuracies can be compared to the HD found in Figure \ref{fig:hellingerdistance}. Part (b) shows a simple example with an HD of 0.694, and two extreme cases with HDs of 0.176. The first example case has a different distribution than those discussed earlier in this work.

The validation accuracy heatmaps show that the network with K-Origins consistently outperforms the one without it. The network without K-Origins struggles most when $\Delta \sigma = 0$, indicating a pure mean shift. The results also suggest that having $L > RLF$ is beneficial, but this is specific to the dataset used in this paper. Just as small, noisy squares are hard to detect, larger noisy squares become easier to detect with such controlled data. 

KRFL14 also makes relatively good predictions when the HD is 0.176 which is an extremely challenging segmentation task for both machines and humans. This demonstrates the effectiveness of intensity quantification for tasks such as object detection.

After adding K-Origins, the accuracy heatmap is almost directly correlated to the class HDs. This is evident by comparing Figure \ref{fig:hellingerdistance} to the KRFL14 accuracy plots in Figure \ref{fig:twoclassless} and Figure \ref{fig:twoclassmore}. As the HD decreases, so does the accuracy, and vice versa. This correlation is not observed in the traditional network without K-Origins.

\subsubsection{Two Target Classes: Tracer Segmentation}\label{sec:mixedtwotarget}

In this section, we segment two identically shaped target classes (both squares) from a background and differentiate these classes from each other, addressing the "Tracer Problem." This involves varying the intensity distributions of the two target classes to make them more or less similar.

\begin{figure}[ht]
    \centering
    \includegraphics[width=\textwidth]{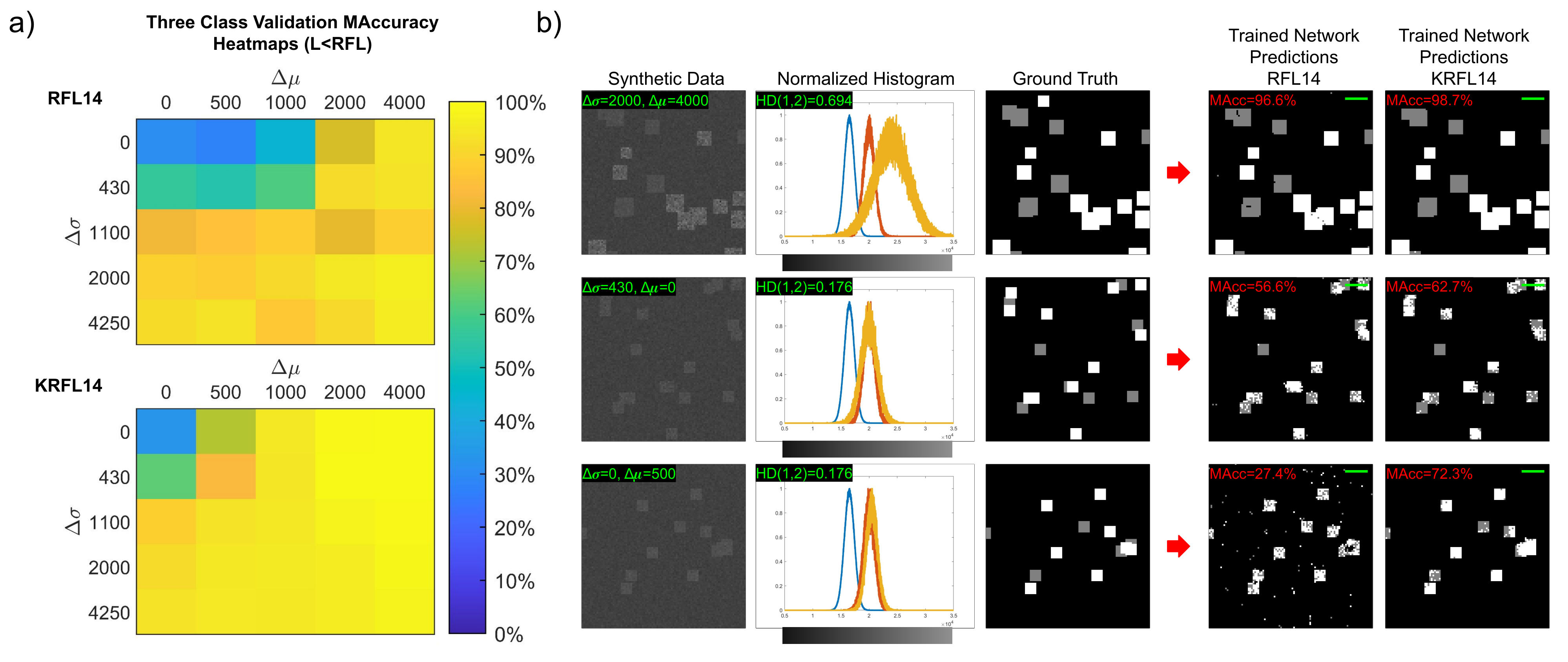}
    \caption[Two target class $L<RFL$ results]{RFL14 and KRFL14 validation results for tracer segmentation with $L<RFL$ synthetic data. (a) Segmentation validation results visualized in a heatmap for various Hellinger distance variations, represented by shifting means and standard deviations. (b) Three example cases: an easy-to-understand one followed by the two hardest cases.}
    \label{fig:threeclassless}
\end{figure}

\begin{figure}[ht]
    \centering
    \includegraphics[width=\textwidth]{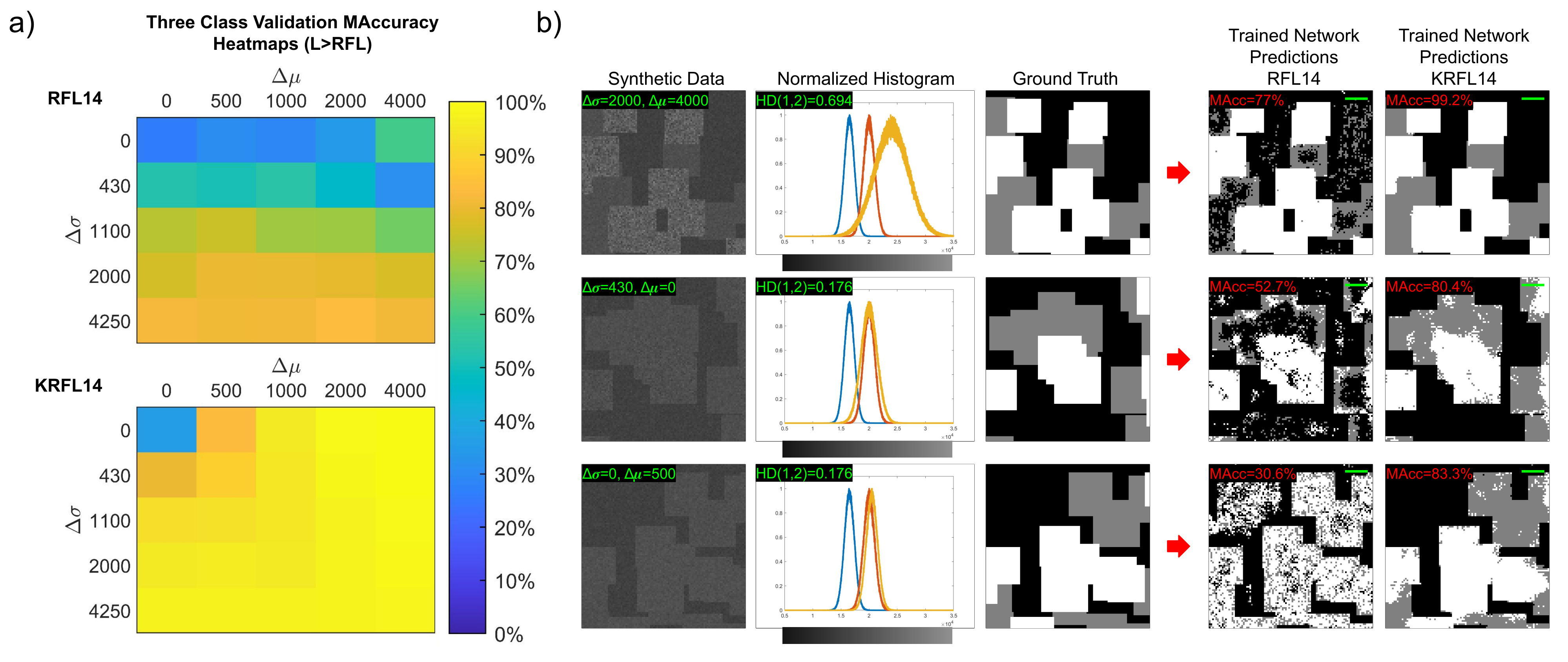}
    \caption[Two target class $L>RFL$ results]{RFL14 and KRFL14 validation results for tracer segmentation with $L>RFL$ synthetic data. (a) Segmentation validation results visualized in a heatmap for various Hellinger distance variations, represented by shifting means and standard deviations. (b) Three example cases: one easy-to-understand example followed by the two hardest cases.}
    \label{fig:threeclassmore}
\end{figure}

The background has $\mu_0 = 16500$ and noise with a standard deviation $\sigma_0 = 900$ to minimize interference with the target classes. The first target class has $\mu_1 = 20000$ and $\sigma_1 = 1000$ , while the second target class varies based on $\Delta \mu = \mu_2-\mu_1$ and $\Delta \sigma = \sigma_2 - \sigma_1$. Results for $L<RFL$ are shown in Figure \ref{fig:threeclassless}, and results for $L>RFL$ are shown in Figure  \ref{fig:threeclassmore}. In part (a), we present the heatmaps showing validation accuracy results for RFL14 and KRFL14, which can be compared to a trials HD using Figure \ref{fig:hellingerdistance}. Part (b) provides a straightforward example followed by the two most challenging cases tested.

KRFL14 consistently outperforms RFL14 in this task, especially when the standard deviation remains constant while $\Delta \mu$ varies. Accuracy increases with $L>RFL$ are for the same reason as mentioned before. This shows extremely promising results, with useful segmentation even at an HD of 0.176. As in Section \ref{sec:mixedonetarget}, the accuracy plots for the network using K-Origins correlates well with the HD plot.

\section{Discussion}

The most significant improvements from K-Origins occur when the primary difference between class intensity distributions is a mean shift. In reality, this scenario is common because classes often differ by colour or intensity means.

The RFL-related experiments suggest that a network's depth should be set such that the RFL is larger than all target object sizes (L$<$RFL). Soon after this point, the primary author hypothesizes that making the architecture deeper is less beneficial and less efficient than making it wider. K-Origins is an example of making it wider, as is increasing the number of filters in layers. This could possibly be presented as a guideline for how deep a neural network should be, and could make the design process more deterministic. There is, however, the possibility that these results stem from using such controlled data. This could be studied in the future. 

A potential use case for data without such well-defined intensity distributions, such as general image data, is to use $N$ equally spaced weights along the entire intensity or colour spectrum in each channel, dividing it into $N+1$ different regions for the network to leverage. This approach would make it easier, for example, to determine if a picture of a dog has white snow or green grass in the background. This would likely also help determine the exact colour of the dog. For this reason, K-Origins is likely useful for other classification problems, not just semantic segmentation. The downside, however, is that as $N$ increases, the memory requirements grow significantly due to the number of image copies being created. There are likely ways to make this more efficient.

Additionally, it is unclear if K-Origin layers after the first impact classification results significantly. These weights likely need to be much smaller than those used in this paper and this could be explored in future studies. There is also the possibility to extend the application of K-Origins to un-supervised problems, perhaps by using a modified version of the simple colour network in Figure \ref{fig:clayer}.

The experiments in this paper did not involve any hyperparameter tuning, which would likely improve results significantly, nor were the networks necessarily trained to steady state. Our goal was to demonstrate that even with minimal tuning, the ability to understand colour magnitudes is beneficial for predictions.

\section{Conclusion}

The experimental results from this study suggest that encoder-decoder networks struggle with classifications that require an understanding of colour or intensity magnitudes, as opposed to gradients alone. The custom layer K-Origins, which can be added to any network, was tested by incorporating it in modified U-Net architectures. By adding K-Origins and ensuring a sufficient RFL, there were significant accuracy improvements for the object detection and tracer segmentation problems. This approach allows for the development of smaller and more efficient networks.

These improvements are likely relevant to many fields, as object detection and tracer segmentation problems are common. Additionally, as new network architectures are being studied it would be valuable to test the impact of K-Origins on these emerging architectures, given its compatibility with any network.

\bibliography{templateArxiv}

\appendix

\section{Code}

The Python (TensorFlow) code used in this paper can be found in the GitHub repository associated with the primary author's thesis: \url{https://github.com/lewismmason/Thesis-Public}.

\section{Figure 1}

The networks used for Figure \ref{fig:exampledata} are RFL32 and KRFL32, deeper versions of RFL14 and KRFL14. The additional level is added the same way RFL8 is extended to RFL18. This depth satisfies RFL requirements, and still demonstrates that adding K-Origins is beneficial.

\section{Experimental Data}\label{app:data}

\begin{table}[H]
    \centering
    \begin{tabular}{c|cccccc}
    
         & \multicolumn{5}{c}{L/RFL} \\
         & 0.3 & 0.6 & 0.95 & 1.3 & 2 & 3 \\
        \hline
        RFL8 & 0.9964 & 0.9923 & 0.9957 & 0.5299 & 0.4516 & 0.3449\\
        KRFL8 & 0.9993 & 0.9987 & 0.9977 & 0.9998 & 0.9997 & 0.9999 \\
        \hline
        RFL18 & 0.9991 & 0.993  & 0.9944 & 0.5442 & 0.3194 & 0.4737  \\
        KRFL18 & 0.9967 & 0.9979 & 0.9990 & 0.9999 & 0.9999 & 0.9999 \\
        \hline
        RFL38 & 0.9973 & 0.9966 & 0.975  & 0.9909 & 0.1292 & 0.2694  \\
        KRFL38 & 0.999  & 0.997  & 0.9999 & 0.9999 & 0.9997 & 0.9996 \\
    \end{tabular}
    \caption[RFL Sweep Without Noise Data]{Figure \ref{fig:rflresults}a data.}
    \label{tab:RFL_sweep_no_noise}
\end{table}

\begin{table}[H]
    \centering
    \begin{tabular}{c|cccccc}
         & \multicolumn{5}{c}{L/RFL} \\
        & 0.3 & 0.6 & 0.95 & 1.3 & 2 & 3 \\
        \hline
        RFL8 & 0.7058 & 0.7933 & 0.6347 & 0.3976 & 0.4495 & 0.4592\\
        KRFL8 & 0.8357 & 0.9285 & 0.9674 & 0.9883 & 0.99   & 0.9913 \\
        \hline
        RFL18 & 0.9254 & 0.9051 & 0.9254 & 0.9058 & 0.8287 & 0.9135 \\
        KRFL18 & 0.9105 & 0.9834 & 0.9904 & 0.9935 & 0.9967 & 0.9966 \\
        \hline
        RFL38 & 0.9557 & 0.986  & 0.9721 & 0.9818 & 0.9839 & 0.9885 \\
        KRFL38 & 0.9745 & 0.991  & 0.9901 & 0.9945 & 0.9948 & 0.9966 \\
    \end{tabular}
    \caption[RFL Sweep With Noise Data]{Figure \ref{fig:rflresults}b data.}
    \label{tab:RFL_sweep_noise}
\end{table}

\begin{table}[H]
    \centering
    \begin{tabular}{cc|ccccc} 
        && \multicolumn{5}{c}{$\Delta \mu$} \\
        && 0 & 500 & 1000 & 2000 & 4000\\
        \hline
        &0 & 0.063 & 0.062 & 0.282 & 0.800 & 0.987 \\
        &430 & 0.229 & 0.179 & 0.421 & 0.628 & 0.945 \\
       $\Delta \sigma$  &1100 & 0.559 & 0.706 & 0.656 & 0.819 & 0.959 \\
        &2000 & 0.714 & 0.841 & 0.824 & 0.799 & 0.960 \\
        &4250 & 0.889 & 0.928 & 0.927 & 0.930 & 0.9311 \\
    \end{tabular}
    \caption[RFL14: One Class L<RFL Results]{Figure \ref{fig:twoclassless}a RFL14 data (One class L<RFL).}
    \label{tab:RFL14_1c_L<RFL}
\end{table}

\begin{table}[H]
    \centering
    \begin{tabular}{cc|ccccc} 
        && \multicolumn{5}{c}{$\Delta \mu$} \\
        && 0 & 500 & 1000 & 2000 & 4000\\
        \hline
        &0 & 0.089 & 0.450 & 0.737 & 0.959 & 0.9981 \\
        &430 & 0.421 & 0.549 & 0.703 & 0.926 & 0.995 \\
        $\Delta \sigma$&1100 & 0.702 & 0.728 & 0.780 & 0.888 & 0.987 \\
        &2000 & 0.800 & 0.811 & 0.837 & 0.897 & 0.975 \\
        &4250 & 0.904 & 0.906 & 0.898 & 0.913 & 0.959 \\
    \end{tabular}
    \caption[KRFL14: One Class L<RFL Results]{Figure \ref{fig:twoclassless}a KRFL14 data (One class L<RFL).}
    \label{tab:KRFL14_1c_L<RFL}
\end{table}

\begin{table}[H]
    \centering
    \begin{tabular}{cc|ccccc} 
        && \multicolumn{5}{c}{$\Delta \mu$} \\
        && 0 & 500 & 1000 & 2000 & 4000\\
        \hline
        &0 & 0.150 & 0.545 & 0.574 & 0.665 & 0.749 \\
        &430 & 0.786 & 0.834 & 0.779 & 0.721 & 0.726 \\
        $\Delta \sigma$&1100 & 0.907 & 0.901 & 0.916 & 0.937 & 0.979 \\
        &2000 & 0.954 & 0.945 & 0.944 & 0.965 & 0.991 \\
        &4250 & 0.970 & 0.977 & 0.976 & 0.982 & 0.985 \\
    \end{tabular}
    \caption[RFL14: One Class L>RFL Results]{Figure \ref{fig:twoclassmore}a RFL14 data (One class L>RFL).}
    \label{tab:RFL14_1c_L>RFL}
\end{table}

\begin{table}[H]
    \centering
    \begin{tabular}{cc|ccccc} 
        && \multicolumn{5}{c}{$\Delta \mu$} \\
        && 0 & 500 & 1000 & 2000 & 4000\\
        \hline
        &0 & 0.129 & 0.837 & 0.937 & 0.991 & 0.999 \\
        &430 & 0.802 & 0.856 & 0.920 & 0.981 & 0.999 \\
        $\Delta \sigma$&1100 & 0.909 & 0.923 & 0.934 & 0.970 & 0.997 \\
        &2000 & 0.945 & 0.956 & 0.958 & 0.970 & 0.993 \\
        &4250 & 0.978 & 0.973 & 0.980 & 0.977 & 0.985 \\
    \end{tabular}
    \caption[KRFL14: One Class L>RFL Results]{Figure \ref{fig:twoclassmore}a KRFL14 data (One class L>RFL).}
    \label{tab:KRFL14_1c_L>RFL}
\end{table}

\begin{table}[H]
    \centering
    \begin{tabular}{cc|ccccc} 
        && \multicolumn{5}{c}{$\Delta \mu$} \\
        && 0 & 500 & 1000 & 2000 & 4000\\
        \hline
        &0 & 0.309 & 0.274 & 0.441 & 0.766 & 0.944 \\
        &430 & 0.566 & 0.530 & 0.606 & 0.915 & 0.936 \\
        $\Delta \sigma$&1100 & 0.816 & 0.851 & 0.881 & 0.794 & 0.885 \\
        &2000 & 0.893 & 0.878 & 0.913 & 0.947 & 0.966 \\
        &4250 & 0.9199 & 0.935 & 0.871 & 0.912 & 0.9594 \\
    \end{tabular}
    \caption[RFL14: Two Class L<RFL Results]{Figure \ref{fig:threeclassless}a RFL14 data (Two class L<RFL).}
    \label{tab:RFL14_2c_L<RFL}
\end{table}

\begin{table}[H]
    \centering
    \begin{tabular}{cc|ccccc} 
        && \multicolumn{5}{c}{$\Delta \mu$} \\
        && 0 & 500 & 1000 & 2000 & 4000\\
        \hline
        &0 & 0.332 & 0.725 & 0.946 & 0.990 & 0.9948 \\
        &430 & 0.627 & 0.825 & 0.942 & 0.995 & 0.995 \\
        $\Delta \sigma$&1100 & 0.887 & 0.936 & 0.949 & 0.977 & 0.991 \\
        &2000 & 0.915 & 0.948 & 0.954 & 0.961 & 0.987 \\
        &4250 & 0.934 & 0.951 & 0.949 & 0.958 & 0.977 \\
    \end{tabular}
    \caption[KRFL14: Two Class L<RFL Results]{Figure \ref{fig:threeclassless}a KRFL14 data (Two class L<RFL).}
    \label{tab:KRFL14_2c_L<RFL}
\end{table}

\begin{table}[H]
    \centering
    \begin{tabular}{cc|ccccc} 
        && \multicolumn{5}{c}{$\Delta \mu$} \\
        && 0 & 500 & 1000 & 2000 & 4000\\
        \hline
        &0 & 0.260 & 0.306 & 0.290 & 0.344 & 0.593 \\
        &430 & 0.527 & 0.510 & 0.537 & 0.467 & 0.313 \\
        $\Delta \sigma$&1100 & 0.728 & 0.753 & 0.702 & 0.697 & 0.652 \\
        &2000 & 0.764 & 0.8044 & 0.800 & 0.796 & 0.770 \\
        &4250 & 0.823 & 0.811 & 0.816 & 0.839 & 0.813 \\
    \end{tabular}
    \caption[RFL14: Two Class L>RFL Results]{Figure \ref{fig:threeclassmore}a RFL14 data (Two class L>RFL).}
    \label{tab:RFL14_2c_L>RFL}
\end{table}

\begin{table}[H]
    \centering
    \begin{tabular}{cc|ccccc} 
        && \multicolumn{5}{c}{$\Delta \mu$} \\
        && 0 & 500 & 1000 & 2000 & 4000\\
        \hline
        &0 & 0.355 & 0.833 & 0.955 & 0.993 & 0.999 \\
        &430 & 0.804 & 0.890 & 0.946 & 0.986 & 0.998 \\
        $\Delta \sigma$&1100 & 0.925 & 0.933 & 0.948 & 0.976 & 0.996 \\
        &2000 & 0.954 & 0.963 & 0.954 & 0.980 & 0.992 \\
        &4250 & 0.974 & 0.973 & 0.974 & 0.978 & 0.982 \\
    \end{tabular}
    \caption[KRFL14: Two Class L>RFL Results]{Figure \ref{fig:threeclassmore}a KRFL14 data (Two class L>RFL).}
    \label{tab:KRFL14_2c_L>RFL}
\end{table}

\end{document}